# Prototype Embedding Optimization for Human-Object Interaction Detection in Livestreaming

PeO-HOI

Menghui Zhang
*School of Information Science and Technology*
*Beijing University of Technology*
Beijing, China
zzmenghui@emails.bjut.edu.cn

Jing Zhang
*School of Information Science and Technology*
*Beijing Key Laboratory of Computational Intelligence and Intelligent System*
*Beijing University of Technology*
Beijing, China
zhj@bjut.edu.cn

Lin Chen
*School of Information Science and Technology*
*Beijing University of Technology*
Beijing, China
cl14cool@emails.bjut.edu.cn

Li Zhuo
*School of Information Science and Technology*
*Beijing Key Laboratory of Computational Intelligence and Intelligent System*
*Beijing University of Technology*
Beijing, China
zhuoli@bjut.edu.cn

***Abstract*—Livestreaming frequently features interactions between streamers and objects, which is critical for understanding and regulating online content. While human-object interaction (HOI) detection has advanced significantly for general video tasks, its application to recognizing streamer-object interactions in livestreaming often exhibits object bias: an excessive focuses on the objects at the expense of interactions with the streamer. To address this challenge, we propose a prototype embedding optimization for human-object interaction detection (PeO-HOI). Our approach first preprocesses the livestreaming using object detection and tracking to extract features of human-object (HO) pairs. Subsequently, prototype embedding optimization is applied to mitigate object bias effects. Finally, after modeling the spatio-temporal context among HO pairs, the HOI detection results are generated by the prediction head. Experimental results demonstrate that the PeO-HOI achieves detection accuracies of 37.19%@full, 51.42%@non-rare, and 26.20%@rare on the publicly available VidHOI dataset, and 45.13%@full, 62.78%@non-rare, and 30.37%@rare on our self-built BJUT-HOI dataset. These results confirm that PeO-HOI effectively enhances HOI detection performance in livestreaming scenarios.**

## I. Introduction

Livestreaming has become an indispensable part of the public's entertainment life. As the content dominator of livestreaming, certain streamer mix in illegal content to gain traffic, which brings great pressure on network regulation. During livestreaming, there are often interactions between the streamer and the objects, such as live commerce and workouts, etc. The human-object interaction (HOI) is crucial for understanding and regulating livestreaming content. HOI detection [1-3] aims to infer the relationships between humans and objects in a video stream, by identifying the objects that are interacting with the humans and estimating the type of interaction. Most of the current state-of-the-art (SOTA) techniques mainly focus on HOI detection in still images and are not yet able to fully address the challenge in videos. They can be broadly categorized into one-stage and two-stage methods. One-stage methods, such as QPIC (Tamura et al. [4]), STTran (Cong et al. [5]), TUTOR (Tu et al. [6]), and TPT (Zhang et al. [7]), are not easy to model the human-object (HO) union regions effectively when spatio-temporal contextual information is more complex, and it is not promising to obtain well generalization results, especially for the dynamic interactive livestreaming with frequent changes. Two-stage models [5,8] are generally divided into region proposal and interaction classification [9]. For example, Ni et al. [10] pioneered the use of human gaze features in HOI prediction and designed a two-stage multimodal framework for HOI detection and prediction in videos with SOTA performance by utilizing visual appearance features, semantic contexts, and human gaze cues, thus becoming a benchmark for many subsequent HOI efforts.

However, existing HOI detection models focus too much on the information objects so as to neglect the interaction relationship between human and objects, leading to the issue of object bias and is not favorable for livestreaming applications. A feasible way to solve the problem is to make the model prioritize HOIs when constructing the feature space. Liu et al. [11] proposed the object class immunity (OC-immunity) network to generalize to unseen objects by decoupling OC-immune representations. Wang et al. [12] proposed the object-wise debiasing memory (ODM) method to achieve verb balancing under each object, and to assign high sampling priority to rare class instances. Driven by their perspectives, we emphasize the importance of modeling HO union features. In contrast, prototype learning [13] is more effective in learning discriminative prototype representations for differentiating categories in the feature space. For example, Li et al. [14] applied prototype learning to obtain more discriminative features using labeled frame features as category prototypes by imposing category relation constraints with contrastive representation learning, and then generating pseudo labels based on the semantic similarity of the features in the embedding space. Given this advantages, we will design a prototype embedding optimization mechanism to generate a compact and unique union feature representation of HO pairs to mitigate object bias in HOI detection.

In this work, we are committed to exploring prototype embedding optimization for HOI detection (PeO-HOI) in livestreaming to improve model accuracy and generalization by mitigating object bias. We introduce a prototype learning pipeline by constructing the HO pair union feature representation to embed interactive feature prototype into the corresponding HO pair features. Subsequently, due to the frequent dynamic changes of HOIs in livestreaming and the difficulty of labeling, there is a label noise problem in the livestreaming data samples. Therefore, we propose the propensity-reweighted cross-entropy loss ($L_{\mathrm{PWCE}}$) to optimize the union feature space of HO pairs by adaptively adjusting the weights of samples. In addition, considering the spatio-

This work in this paper was supported in part by the National Natural Science Foundation of China under Grant 62471013, Grant 61971016.

temporal characteristics of livestreaming, we also design a spatio-temporal relationship module that aggregate contexts in a sliding window frames to improve HOI detection performance.

Our contribution can be mainly summarized as follows:

- We introduce a streaming-aware prototype learning pipeline that embeds interactive feature prototypes to generate HOI union feature representations, mitigating spatial bias in dynamic objects for livestreaming HOI.
- We propose a propensity-reweighted cross-entropy loss to optimize the HO pair union feature space, effectively addressing label noise problem in livestreaming HOI.
- We design a sliding window spatio-temporal relationship module to fuse temporally evolving HOI pair features within livestreaming contexts.

## II. Method

Fig.1 illustrates the overall structure of our prototype embedding optimization for human-object interaction (PeO-HOI) detection in livestreaming. It consists of three parts (a) Feature extraction of HO pairs: we extracting the HO features, word embedding features, then match them to form HO candidate pairs $\{\mathbf{H}_j, \mathbf{O}_i\}$ and their union features; (b) Prototype embedding optimization: to address object bias in HOI detection, we embed semantic prototype as prior knowledge to guide the generation of HO pair union feature space, and then $L_{\mathrm{PWCE}}$ is applied to optimize the union feature space by adaptively adjusting the sample weights to enhance the HO pair feature $\mathbf{HO}_t$; (c) HOI detection: the embedded HO pair features are fed into the spatial attention encoder to to build global context representations $\mathbf{HO}^{\mathrm{S}}$, which are then fused into the spatio-temporal features $\mathbf{HO}^{\mathrm{ST}}$ of the HOs, after which the spatial interaction and action interaction prediction heads are used to predict HOI detection results, respectively.

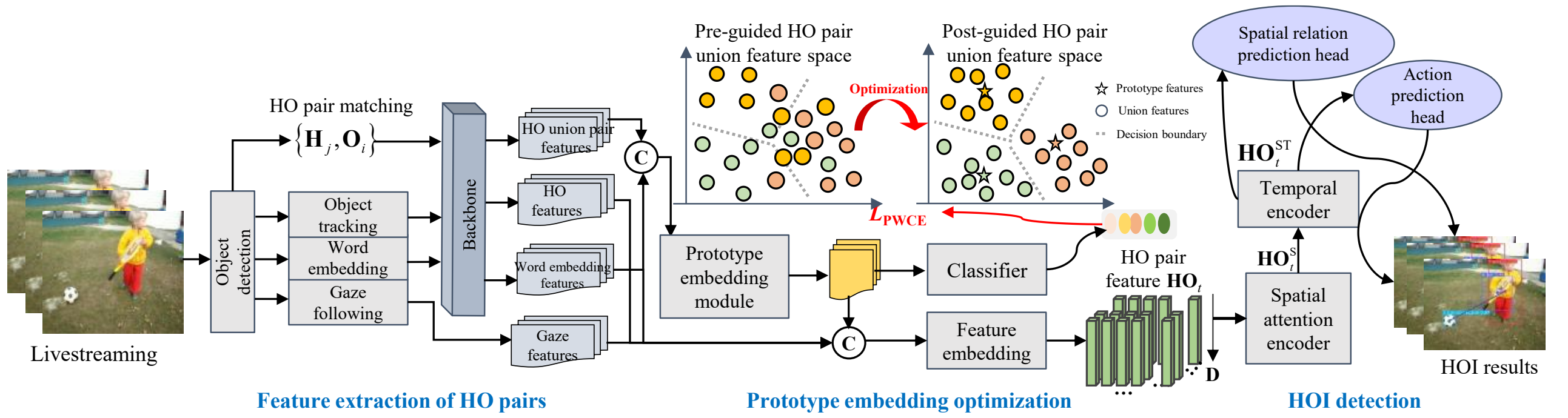

Fig. 1. The proposed PeO-HOI detection.

### A. Feature Extraction of Human-Object Pairs

In this paper, we model the HO pair feature representation from multiple dimensions such as visual appearance, word features, and streamer's gaze cues. We adopt an interactive system developed by Ni et al. [10] to produce the HO pair representation, including YOLOv5 [15] for object detection, DeepSort [16] for tracking, GloVe [17] for word embedding features such as the visual appearance of streamers and objects, word features, and the streamer's gaze cues. In detail, the feature extraction process is carried out using $\Phi^{\mathrm{det}}$ (YOLOv5), $\Phi^{\mathrm{tra}}$ (DeepSort), $\Phi^{\mathrm{va}}$ (ResNet-101), $\Phi^{\mathrm{w}}$ (GloVe), and $\Phi^{\mathrm{g}}$.

(1) The video frame sequence $\mathbf{I}$ is fed into $\Phi^{\mathrm{det}}$ and the pre-trained $\Phi^{\mathrm{tra}}$ to detect the candidate boxes containing streamer and objects in the keyframes to obtain the object position and trajectory. This helps obtain the streamer and object frames $\mathbf{b}^{\mathrm{h}}_{t,j}$ and $\mathbf{b}^{\mathrm{o}}_{t,i}$, object categoriy $\mathbf{c}_{t,i}$, and the motion trajectory features $\mathbf{H}_j$ and $\mathbf{O}_i$ for both the streamer and the objects.

(2) Visual appearance features $\Phi^{\mathrm{va}}$ is used to extract the streamer and object features $\mathbf{v}^{\mathrm{h}}_{t,j}$, $\mathbf{v}^{\mathrm{o}}_{t,i}$=$\Phi^{\mathrm{va}}$($\mathbf{b}^{\mathrm{h}}_{t,j}$, $\mathrm{b}^{\mathrm{o}}_{t,i}$).The predicate relationship between streamer $\mathbf{b}^{h}_{t,j}$ and object $\mathbf{b}^{\mathrm{o}}_{t,i}$ are represented by the union features $\mathbf{v}^{\mathrm{u}}_{t,<i,j>}$ that is derived from the union frame $\mathbf{b}^{\mathrm{u}}_{t,<i,j>}$: $\mathbf{v}^{\mathrm{u}}_{t,<i,j>}$=$\Phi^{\mathrm{va}}$($\mathbf{b}^{\mathrm{u}}_{t,<i,j>}$).

(3) Word embedding features $\Phi^{\mathrm{w}}$ is used to generate category-specific word vectors $\mathbf{w}_{t,i}$ for object categories $\mathbf{c}_{t,i}$ to distinguish different categories and simplify subsequent processing: $\mathbf{w}_{t,i}$=$\Phi^{\mathrm{w}}$($\mathbf{c}_{t,i}$).

(4) Gaze cues $\Phi^{\mathrm{g}}$ is introduced to extract the prior knowledge $\mathbf{g}_{t,j}$ of gaze direction from the streamer frame $\mathbf{b}^{\mathrm{h}}_{t,j}$: $\mathbf{g}_{t,j}$=$\Phi^{\mathrm{g}}$($\mathbf{b}^{\mathrm{h}}_{t,j}$).

(5) The HO pair feature $\mathbf{HO}_t$ is constructed by combining the above visual appearance features, word vectors and streamer gaze cues.

$$\mathbf{HO}_t = \mathrm{FC}\left(\mathrm{Concat}\left(\mathbf{v}^{\mathrm{h}}_{t,j}, \mathbf{v}^{\mathrm{o}}_{t,i}, \mathbf{v}^{\mathrm{u}}_{t,<i,j>}, \mathbf{w}_{t,i}, \mathbf{g}_{t,j}\right)\right). \quad (1)$$

### B. Prototype Embedding Optimization

Prototype learning follows the human learning way by creating clear semantic associations in the feature space through the representative feature prototypes of different interactions, resulting in stronger feature representations. Based on it, we design a prototype embedding optimization mechanism to optimize the HO pair features using shared semantic representation of word embeddings to solve the issue of feature deviation in livestreaming due to big differences in object appearance and uneven frequency distribution. Furthermore, we introduce the propensity score in the loss function to reduce the noisy or unimportant samples to diminish their impact on model training.

As shown in Fig. 2, the proposed prototype embedding optimization mechanism includes the prototype embedding module and a propensity-reweighted cross-entropy loss $L_{\mathrm{PWCE}}$. In this paper, word embedding features are embedded into the feature space of the HO union pair as the prototype prior knowledge, and then the feature space is optimized by connecting a classifier using $L_{\mathrm{PWCE}}$ .

*1) Prototype embedding module:* Using semantic embedding vector to map the discrete features of HO pairs into a semantic subspace centered on the interactive relationship, the model can get rid of the over-reliance on object category features, and instead capture more generalized spatio-temporal interaction patterns to produce more robust HOI feature representations.

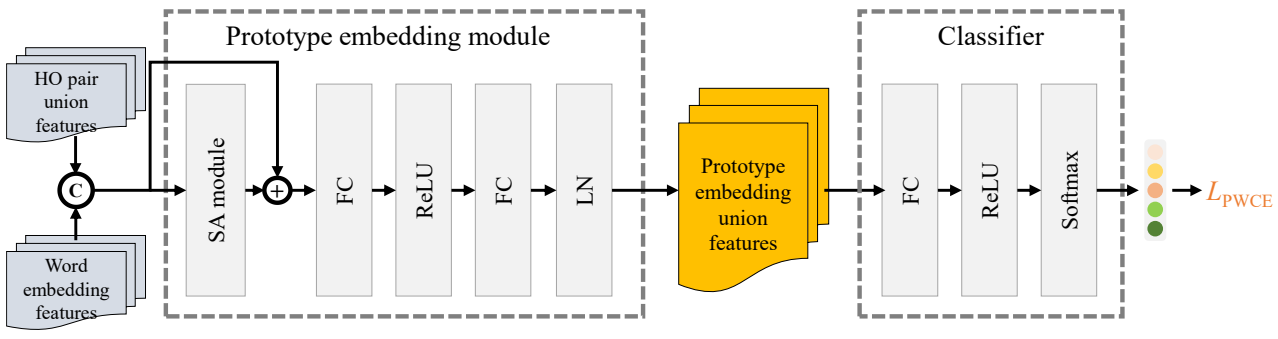

Fig. 2. Prototype embedding optimization mechanism.

The steps are as follows:

a) Concate the matched HO pair union feature $\mathbf{v}^{\mathrm{u}}_{t,<i,j>}$ with the corresponding interaction predicate embedding feature $\mathbf{s}_{t,i}$ to obtain the concated feature $\mathbf{F}_t$:

$$\mathbf{F}_t = \mathrm{Concat}\left(\mathbf{v}^{\mathrm{u}}_{t,<i,j>}, \mathbf{s}_{t,i}\right). \tag{2}$$

b) Feed $\mathbf{F}_t$ into a self-attention module with residual structure to capture the correlation between features to obtain the correlated features $\mathbf{F}_t'$:

$$\mathbf{Q}_t = f_{\mathrm{Q}}\left(\mathbf{F}_t\right), \mathbf{K}_t = f_{\mathrm{K}}\left(\mathbf{F}_t\right), \mathbf{V}_t = f_{\mathrm{V}}\left(\mathbf{F}_t\right), \tag{3}$$

$$\begin{aligned} \mathbf{F}_t' &= \mathbf{F}_t + \mathrm{Attention}(\mathbf{Q}_t, \mathbf{K}_t, \mathbf{V}_t) \\ &= \mathbf{F}_t + \mathrm{Softmax}\left(\mathbf{Q}_t \cdot \mathbf{K}_t^{\mathrm{T}} \big/ \sqrt{\mathbf{D}}\right)\mathbf{V}_t \end{aligned}, \tag{4}$$

where $f_{\mathrm{Q}}$, $f_{\mathrm{K}}$, $f_{\mathrm{V}}$ denote the weight matrices of query, key, and value, respectively; $\mathbf{Q}_t$, $\mathbf{K}_t$, $\mathbf{V}_t$ denote the query, key, and value of the HO union features, respectively; and $\mathbf{D}$ is the size of the feature dimension.

c) Feed $\mathbf{F}_t'$ into a feedforward network consisting of a fully connected layer FC, a RELU activation function, and an LN regularization layer to obtain the prototype embedded HO pair union features $\mathbf{F}_t^{\mathrm{pro_u}}$:

$$\mathbf{F}_t^{\mathrm{pro_u}} = \mathrm{LN}\left(\mathrm{FC}\left(\mathrm{RELU}\left(\mathrm{FC}\left(\mathbf{F}_t'\right)\right)\right)\right). \tag{5}$$

*2) Propensity-reweighted cross-entropy loss:* $L_{\mathrm{PWCE}}$ is obtained in the cross-entropy function by introducing a propensity score to dynamically adjust the weights of the samples according to their features and importance, so that the noisy or unimportant smaples will have lower weights , thus migrating the impact on model training. In order to better optimize the HO pair union feature space in the case of label noise, we input the prototype embedded HO pair union feature $\mathbf{F}_{\mathrm{t}}^{\mathrm{pro_u}}$ into the classifier consisting of FC, ReLU and softmax, and introduce the $L_{\mathrm{PWCE}}$ to promote the model's clustering of samples in the feature space with similar interaction semantics, so as to strengthen the discriminative power of the features.

a) Feed the prototype embedding union feature $\mathbf{F}_t^{\mathrm{pro_u}}$ into the classifier to obtain the $\mathrm{P}_u$ according to the HOI labels.

$$\mathrm{P}_u = \mathrm{softmax}\left(\mathrm{RELU}\left(\mathrm{FC}\left(\mathbf{F}_t^{\mathrm{pro_u}}\right)\right)\right). \tag{6}$$

b) To deal with the object bias in the feature space more efficiently, we introduce propensity-reweighted cross-entropy. The propensity score $\alpha_l$ for each label is defined as [18]:

$$\alpha_l = \frac{1}{1 + C\exp - \log(N_l)}, \tag{7}$$

where $N_l$ is the frequency of label $l$ and $C$=(log$N_l$–1) is the weighting coefficient.

c) After inverting the propensity score $\alpha_l$ to obtain the label weight $\omega_l$, it is summed with the cross-entropy loss to obtain the $L_{\mathrm{PWCE}}$:

$$L_{\mathrm{PWCE}} = \sum_l \omega_l \sum_i p_l^i \log\left(p_l^i\right) + \left(1 - p_l^i\right)\log\left(1 - p_l^i\right). \tag{8}$$

d) Fuse the extracted streamer features $\mathbf{v}^h_{t,j}$, object features $\mathbf{v}^{\mathrm{o}}_{t,i}$ , the prototype embedding optimized HO pair union feature $\mathbf{F}_t^{\mathrm{pro_u}}$, word embedding features $\mathbf{w}_{t,i}$ and streamer gaze cue $\mathbf{g}_{t,j}$ into the FC layer to obtain $\mathbf{HO}_t'$:

$$\mathbf{HO}_t' = \mathrm{FC}\left(\mathrm{Concat}\left(\mathbf{v}^{\mathrm{h}}_{t,j}, \mathbf{v}^{\mathrm{o}}_{t,i}, \mathbf{F}_t^{\mathrm{pro_u}}, \mathbf{w}_{t,i}, \mathbf{g}_{t,j}\right)\right). \tag{9}$$

### *C. HOI Detection in Livestreaming*

The feature representation of HO pairs captures some interaction information, but spatial relationships (such as position, distance, and orientation) and temporal dynamics can further refine these representations. Thus, we use the spatio-temporal relationship module for HOI detection. First, spatial attention encoder (see Algorithm 1) is used to enhance intra-frame relationship modeling by learning feature attention weights, then a temporal window encoder (see Algorithm 2) with cross-attention mechanism is used to dynamically fuse the features in the local temporal segments and global context to produce interactive representations with spatio-temporal consistency.

Algorithm 1: Spatial attention encoder

**Input:** Intra-frame HO relationship $\mathbf{HO}_t$=[$\mathbf{HO}_{t,<1,1>}$,…, $\mathbf{HO}_{t,<i,j>}$]
**Output:** HO relationship $\mathbf{HO}_t^{\mathrm{S}}$, global feature vector $\mathbf{c}_t$
1: **for all** $\mathbf{HO}_{t,<i,j>} \in \mathbf{HO}_t$ **do**
2: Attach a learnable global token $\boldsymbol{\tau}$ to $\mathbf{HO}_{t,<i,j>}$
3: **end for**
4: Apply multi-head self-attention with residual structure to obtain $\mathbf{HO}_t'$
5: Refine through a regularization layer (LN) and a feedforward network (FFN) with residual structure:
$\mathbf{HO}_{\mathrm{t}}^{\mathrm{S}}$, $\mathbf{c}_t$=LN(LN($\mathbf{HO}_t'$)+FFN(LN($\mathbf{HO}_t'$)))
6: **R**eturn $\mathbf{HO}_{\mathrm{t}}^{\mathrm{S}}$, $\mathbf{c}_t$

Algorithm 2: Temporal window encoder

**Input:** $\mathbf{HO}_t^{\mathrm{S}}$, global feature vector $\mathbf{c}_t$, gaze features $\mathbf{g}_{t,j}$
**Output:** HO pair representation $\mathbf{HO}_t^{\mathrm{ST}}$, HOI detection result $\mathrm{P}_t$
1: Refine $\mathbf{HO}_{\mathrm{t}}^{\mathrm{S}}$ through a sliding window to obtain $\mathbf{HO}_t^{\mathrm{S'}}$
2: Add Sin position encoding $\mathrm{PE}_{\sin}(pos)$ :
$\mathbf{HO}_t^{\mathrm{S''}}$ =$\mathbf{HO}_t^{\mathrm{S'}}$ + $\mathrm{PE}_{\sin}(pos)$
where $\mathrm{PE}_{\sin}(pos,i)=\sin(pos/1000^{i/d_{\mathrm{model}}})$
3: Concate gaze features $\mathbf{g}_{t,j}$ as well as the global feature vector $\mathbf{c}_t$ and feed them with $\mathbf{HO}_t^{\mathrm{sp''}}$
4: Apply multi-head self-attention layer and regularization.
5: Fuse via cross-attention and FFN:
$\mathbf{HO}_t^{\mathrm{ST}}$=LN(LN($\mathrm{c}_{\mathrm{t}}^{'''}$)+FFN($\mathrm{c}_{\mathrm{t}}^{'''}$))
6: Compute predictions via heads:
$\mathrm{P}_t$=Concat($f_{\mathrm{s}}$($\mathbf{HO}_t^{\mathrm{ST}}$),$f_{\mathrm{a}}$($\mathbf{HO}_t^{\mathrm{ST}}$))
7: **Return** $\mathbf{HO}_t^{\mathrm{ST}}$, $\mathrm{P}_t$

Since the HO pairs in the HOI belong to a multi-class multi-label classification problem. Furthermore, the dataset exhibits a long-tailed interaction distribution, leading to class imbalance. To address the imbalance issue and avoid overemphasizing the importance of the most frequent classes in the dataset, we employ the class-balanced focal loss $L_{\text{focal}}$:

$$L_{\text{focal}} = \sum_i \frac{1-\beta}{1-\beta^{n_i}} \left(1 - p_{y_i}\right)^{\gamma} \log\left(p_{y_i}\right), \quad (10)$$

where $p_{y\,i}$ is the $i$-th class and the estimated probability $y$, the variable $n_i$ is the number of samples in the ground truth of the $i$-th class, and $\beta$ is an adjustable parameter.

In the training phase, we perform a weighted summation of the class-balanced loss $L_{\text{focal}}$ and the propensity-reweighted cross-entropy loss $L_{\text{PWCE}}$.

$$L = L_{\text{focal}} + \lambda L_{\text{PWCE}}, \quad (11)$$

where $\lambda$ balances the hyperparameters.

## III. EXPERIMENTS

### A. Dataset

To demonstrate the generalization ability of our PeO-HOI, we choose the publicly available VidHOI dataset, which covers various HOI categories in complex video scenarios. In addition, we supplemented the self-built dataset BJUT-HOI to bridge the gap of the existing datasets for livestreaming HOI, so as to improve the accuracy and robustness of our PeO-HOI.

**VidHOI** [8] dataset is currently the largest video dataset with complete HOI annotations, and its application is based on keyframe annotations sampled at 1 frame per second (FPS), with 78 object categories and 50 predicate classes. Among the predicate class, 8 predicates are defined as spatial relations (e.g., next to, behind, etc.) and the remaining 42 predicates are defined as action relations (e.g., push, lift, etc.). The VidHOI dataset represent real-world scenarios and contain 6366 and 756 untrimmed videos for training and testing, respectively.

**BJUT-HOI** dataset removes redundant video samples from VidHOI that do not contain HOI and is extended by collecting data from the livestreaming platform, with a total of 59 object categories and 50 predicate categories containing 3738 and 352 untrimmed videos for training and testing.

### B. Experiment Details

To ensure the fairness in the comparison, our object module follows the existing methods [5,7,8,10,19] using the same YOLOv5 model as the object detector with weighted pre-trained on the COCO dataset and fine-tuned on the VidHOI dataset. Similarly, we use the pre-trained DeepSORT model as a body tracker, ResNet-101 as the feature backbone, and the GloVe model for word embedding. In the gaze tracking module, we also applies YOLOv5 to detect the head from RGB frames, and pre-train the model on the Crowdhuman dataset [20] and VideoAttentionTarget dataset [21]. During the training process, all the weights of the object and gaze modules are frozen. In this paper, we design five sets of experiments to comprehensively evaluate our PeO-HOI method.

### C. Comparison with SOTAs

To validate the performance of the proposed PeO-HOI on VidHOI and BJUT-HOI datasets, we compare with the SOTAs including QPIC [4], TUTOR [6] and STTran [5], which only use visual features, ST-HOI [8], STTran [5], SERVO-HOI [19], AcoLP [22], Ni et al. [10], TPT [7] based on motion trajectory, SERVO-HOI [19] based on human posture, and AcoLP [22], Ni et al. [10] based on gaze. The method of Ni et al. [10] is the main comparison in this study and one of the most advanced for video HOI detection.

**The results on the VidHOI dataset** are shown in TABLE I. The detection accuracy of our PeO-HOI method is 37.19%@full, 51.42%@non-rare, 26.20%@rare, which is 1.55%, 2.11% and 1.06% better than that of the Ni et al method. This indicates that the prototype embedding optimization in our method can significantly improve the discriminative ability of HO union features and enhance the detection of various interaction relationships.

TABLE I. COMPARISON WITH OTHER METHODS ON VIDHOI

| **Method** | **Features** | | | | **mAP (%)** | | |
|---|---|---|---|---|---|---|---|
| | ***A*** | ***T*** | ***P*** | ***G*** | ***Full*** | ***Non-rare*** | ***Rare*** |
| QPIC [4] | √ | | | | 21.40 | 32.90 | 20.56 |
| TUTOR [6] | √ | | | | 26.92 | 37.12 | 23.49 |
| STTran [5] | √ | | | | 28.32 | 42.08 | 17.74 |
| TPT [7] | √ | √ | | | 31.40 | 44.30 | 21.50 |
| ST-HOI [8] | √ | √ | | | 17.60 | 27.20 | 17.30 |
| SERVO-HOI [19] | √ | √ | √ | | 21.10 | 29.20 | 19.50 |
| Ni et al. [10] | √ | √ | | √ | <u>35.64</u> | <u>49.31</u> | 25.14 |
| ACoLP [22] | √ | √ | | | 28.27 | 39.66 | **26.63** |
| PeO-HOI | √ | √ | | √ | **37.19** | **51.42** | <u>26.20</u> |

a. Note: A, T, P and G represent Appearance, Trajectory, Pose and Gaze respectively. Bold indicates optimal and underlined indicates suboptimal.

**The results on BJUT-HOI dataset are shown** in TABLE II. The detection accuracies of our PeO-HOI is 45.13%@full, 62.78%@non-rare, 30.37%@rare, which are 1.60%, 3.17% and 0.28% better than the method of Ni et al. The param and FLOPs are 234.65M and 0.376G, higher 73.11M and 0.003G than Ni et al.'s method. This suggests that our prototype embedding optimization can significantly improve the discriminative properties of HO pair union features and comprehensively enhance the detection of various types of interaction relationships.

TABLE II. COMPARISON WITH OTHER METHODS ON BJUT-HOI

| **Method** | **Map (%)** | | | **Params (M)** | **FLOPs (G)** |
|---|---|---|---|---|---|
| | ***Full*** | ***Non-rare*** | ***Rare*** | | |
| ST-HOI [8] | 27.47 | 39.01 | 27.32 | 223.37 | 0.327 |
| QPIC [4] | 31.58 | 40.06 | 22.64 | 102.74 | 0.012 |
| STTran [5] | 35.27 | 51.86 | 20.37 | 148.75 | 0.018 |
| Ni et al. [10] | <u>43.53</u> | <u>59.61</u> | 30.09 | 170.54 | 0.373 |
| ACoLP [22] | 35.31 | 49.98 | **30.83** | 243.74 | 0.416 |
| PeO-HOI | **45.13** | **62.78** | <u>30.37</u> | 234.65 | 0.376 |

b. Bold indicates optimal and underlined indicates suboptimal.

### D. Visual Results

Fig. 3 exhibits the visual HOI detection results on BJUT-HOI dataset. The video samples from top to bottom are the original video frame image sequences, interaction action labels, and HOI detection results of the proposed PeO-HOI in livestreaming. The correct, failure cases, and missing labels are highlighted with green, red, and blue boxes, respectively. The interaction action label in the second row and the predicted interaction result in the last row are presented as text.

**Correct cases.** In Fig. 3(a), the video labels are rather coarse with labeling noise, e.g., the "away" action in frame 1 does not appear. Nonetheless, the model performs well on the "next to" interaction relation, and successfully detects the action and spatial relation "holding the cup" in frames 2-4, demonstrating the ability of HOI relationship detection in datasets with label noise. Fig. 3(b) shows a video with a complex background where spatial and action interactions are labeled. The model successfully detects the spatial relationship between the streamer and the objects, and accurately identifies the "hold" interaction, which shows the effectiveness of the method in video HOI detection.

**Failure cases.** In Fig. 3(a), except for the label noise in frame 1, the model misrecognizes the tuner as a piano in frame 3, but the interaction detection result is still correct, indicating that the model can accurately detect the interaction regardless of the object category. In Fig. 3(b), the model fails to correctly detect the position correctly in the first stage for small objects such as handle toys, highlighting the limitations of small object interaction processing. Visual results show that despite the presence of label noise, our PeO-HOI can still learn interactive knowledge and mitigate the object bias issue, but there are still challenges with omissions such as occlusion, re-identification, and small objects, etc.

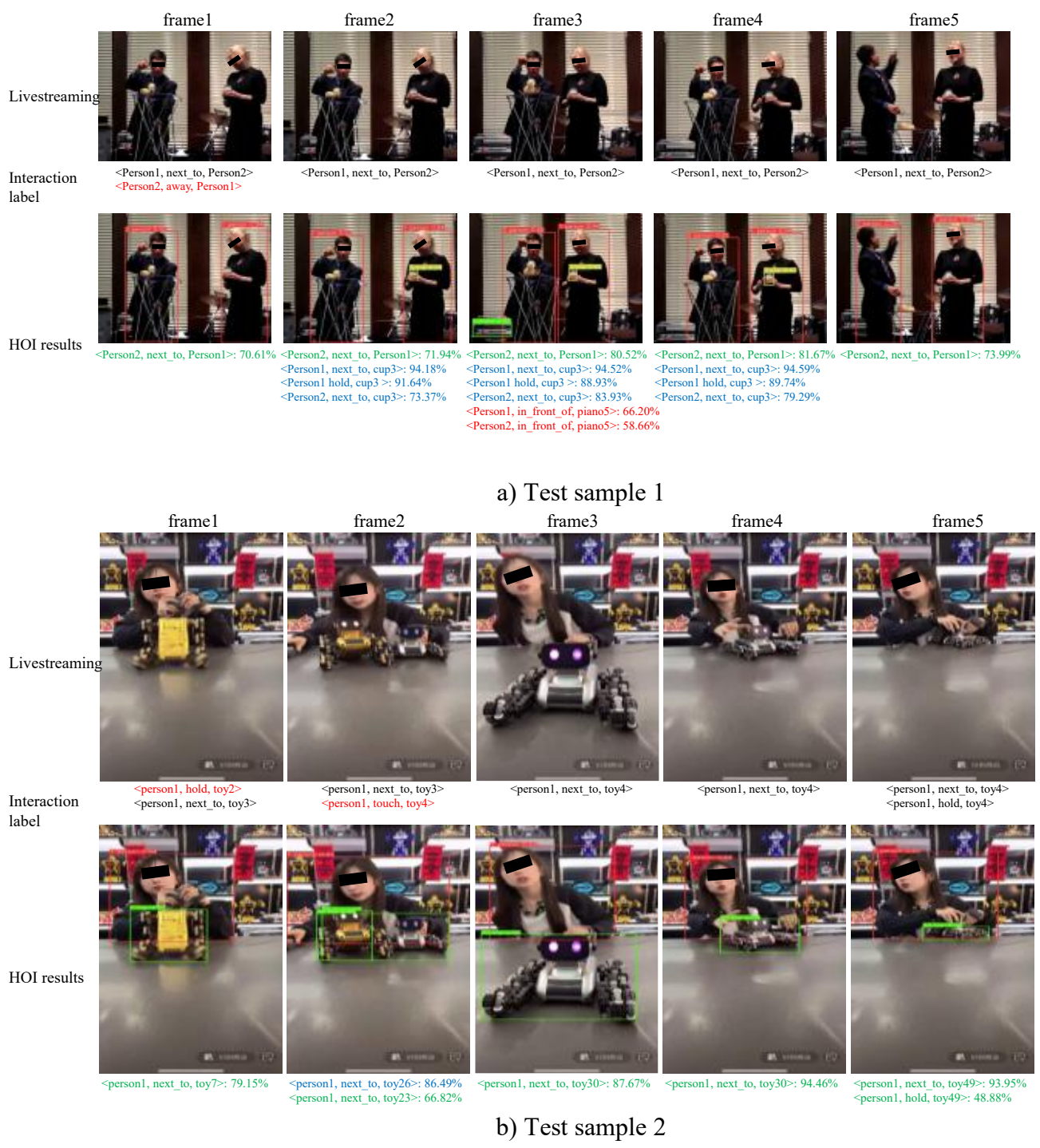

Fig. 3. Visual results on BJUT-HOI dataset.

## E. t-SNE

To demonstrate that prototype embedding union features are more discriminative for the HO pairs, we apply t-SNE on the VidHOI and BJUT-HOI datasets before and after prototype embedding optimization, as shown in Fig. 4. We sample the union feature instance classes that occur more than 400 times in the VidHOI dataset and reserve 400 instance features for visualization. For the BJUT-HOI dataset, the threshold is set to 200. Moreover, instance features regarded as cluster noise are eliminated. Finally, we categorize interaction classes into the top and bottom 25 based on the order of tag IDs, and marked three representative clusters in the t-SNE plots.

Fig. 4 (a), (b) illustrate the feature distributions on the VidHOI via t-SNE projections before using our method. These visualizations indicate that features from different classes overlap with one another, while features belonging to the same class are scattered, suggesting potential limitations in HOI detection. A comparable projection pattern is also observed in the BJUT-HOI dataset, as in Fig. 4 (c). In contrast, Fig. 4 (d-f) visually suggest that our method significantly mitigates the aggregation of interactive features from different classes into the same cluster. As supported by TABLE II, these observations indicate that introduction of prototype embedding optimization can provide more discriminative feature representations and reduce HOI detection errors.

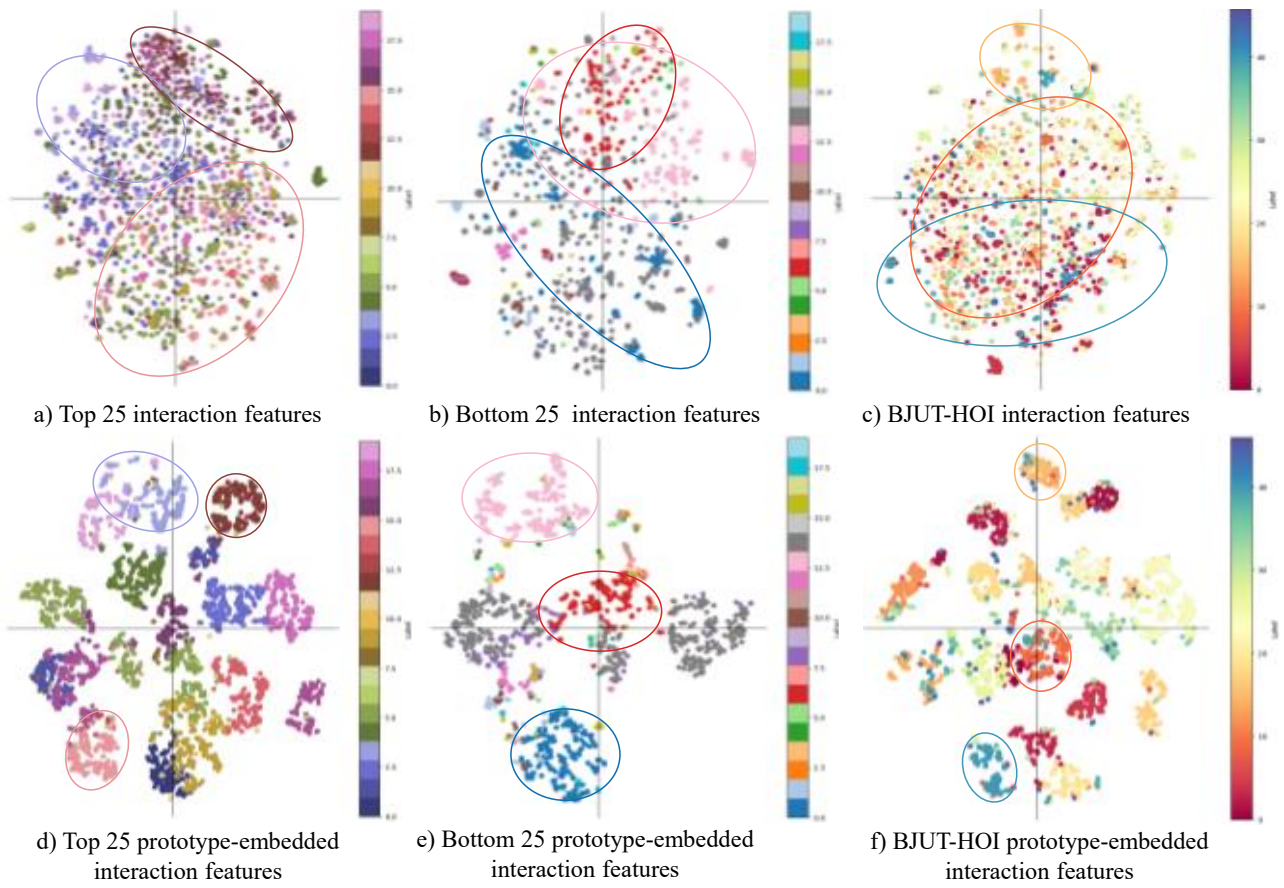

Fig. 4. t-SNE results on feature space.

## F. Parameter Sensitivity Analysis

To investigate the sensitivity of different hyperparameter on model performance, we perform a parameter sensitivity analysis on the BJUT-HOI dataset. TABLE III respectively investigates the influence of hyperparameter $\lambda$ weight between 0.5 and 1.5 on the model performance. We can see the choice of balance coefficient $\lambda$ affects the model's performance, particularly for non-rare classes (mAP@Non-rare), where the variance is 0.136. In contrast, the variance for rare classes (mAP@Rare) is only 0.063. The results show that the quality of HOI union features has a large impact on rare category detection, which confirms the importance of removing object bias and the effectiveness of prototype embedding optimization. When $\lambda$=0.8, the model reaches 45.13%, 62.78% and 30.37% in mAP@Full, mAP@Non-rare and mAP@Rare. Therefore, we follows these hyperparameter setting.

TABLE III. THE INFLUENCE OF HYPERPARAMETER Λ

| $\lambda$ | mAP(%) | | | VAR | | |
|---|---|---|---|---|---|---|
| | *Full* | *Non-rare* | *Rare* | *Full* | *Non-rare* | *Rare* |
| 0.5 | 44.72 | 62.48 | 29.88 | 0.086 | 0.136 | 0.063 |
| 0.8 | **45.13** | **62.78** | **30.37** | | | |
| 1 | 44.63 | 62.26 | 29.90 | | | |
| 1.5 | 44.43 | 61.90 | 29.83 | | | |

## G. Ablation Study

To verify the effect of the prototype embedding optimization, we design a set of ablation studies on BJUT-

HOI dataset. As shown in TABLE IV, group 0 is the baseline model, group 1 combines prototype embedding module (PEN), and group 2 further adds $L_{PWCE}$. Compared with the baseline, group 1 only improves by 0.20% in the non-rare category mAP@Non-rare. Drops of 0.08% and 0.25% in mAP@Full and mAP@Rare, respectively, negatively affecting the rare interaction category. This may be due to the fact that the update gradient of the parameters of the prototype embedding module will become small during training before feature embedding, leading to module deactivation or even false guidance. In group 2, the intuition of our PeO-HOI is to introduce $L_{PWCE}$ that is directly optimized for the PEN to activate the its role, achieving the best performance in mAP@Full, mAP@Non-rare, and mAP@Rare, which are increased by 1.60%, 3.17%, and 0.28%, respectively, compared with group 0. The above results show that the proposed method effectively optimizes the distribution space of HOI features with object bias by using prototype features, thereby improving the performance of video HOI detection.

TABLE IV. ABLATION STUDAY

| Group | Baseline | PEN | $L_{PWCE}$ | mAP | | |
|---|---|---|---|---|---|---|
| | | | | *Full* | *Non-rare* | *Rare* |
| 0 | √ | | | 43.53 | 59.61 | 30.09 |
| 1 | √ | √ | | 43.45 | 59.81 | 29.84 |
| 2 | √ | √ | √ | **45.13** | **62.78** | **30.37** |

## IV. CONCLUSION

In this paper, we propose PeO-HOI, a novel method for HOI detection in livestreaming. First, we preprocess livestreams using object detection and tracking techniques to extract features of HO pairs. Subsequently, we introduce a prototype embedding optimization module to mitigate object bias. Additionally, we model spatio-temporal contextual relationships between HO pairs before generating final HOI detection results through the prediction head. Experiments demonstrate that PeO-HOI achieves competitive performance on both the VidHOI benchmark (37.19% full / 51.42% non-rare / 26.20% rare) and our self-built BJUT-HOI dataset (45.13% full / 62.78% non-rare / 30.37% rare).

While our PeO-HOI obtains promising performance, building prototypes directly using semantic embedding features may lead to inherent limitations in prototype representation completeness and model generalization capability. To address these challenges, future work can focus on decoupling and reconstructing HOI representations specifically for dynamic livestreaming scenarios. This decoupling can be achieved by systematically integrating object functional representations with new object instances extracted from diverse video samples, particularly those from livestreaming scenarios, which in turn can be used to reconstruct more comprehensive prototypes of HOI union feature space. This approach is expected to further enhance the prototype representations in live environments while improving the cross-domain generalization performance through real-time adaptive combinations of rich feature space.